\documentclass[journal,twocolumn,letterpaper]{IEEEJ-MASS}
\ifCLASSINFOpdf
\else
\fi

\usepackage{times,amsmath,epsfig}
\usepackage{fancyhdr}
\usepackage{amsmath}
\usepackage{amsfonts}
\usepackage{amssymb}
\usepackage[utf8]{inputenc}
\usepackage{array}
\usepackage{graphicx}
\usepackage{subfigure}
\usepackage{bm}
\usepackage{breqn}
\usepackage{xcolor}
\usepackage{soul}
\usepackage{amssymb}
\usepackage{color}
\usepackage{algorithm}
\usepackage{algorithmic}
\usepackage[numbers]{natbib}

\usepackage{pifont}
\usepackage[colorlinks,linkcolor=blue]{hyperref}

\usepackage{multirow}
\usepackage{booktabs} 
\usepackage{arydshln}

\makeatletter
\def\UrlAlphabet{%
      \do\a\do\b\do\c\do\d\do\e\do\f\do\g\do\h\do\i\do\j%
      \do\k\do\l\do\m\do\n\do\o\do\p\do\q\do\r\do\s\do\t%
      \do\u\do\v\do\w\do\x\do\y\do\z\do\A\do\B\do\C\do\D%
      \do\E\do\F\do\G\do\H\do\I\do\J\do\K\do\L\do\M\do\N%
      \do\O\do\P\do\Q\do\R\do\S\do\T\do\U\do\V\do\W\do\X%
      \do\Y\do\Z}
\def\UrlDigits{\do\1\do\2\do\3\do\4\do\5\do\6\do\7\do\8\do\9\do\0}
\g@addto@macro{\UrlBreaks}{\UrlOrds}
\g@addto@macro{\UrlBreaks}{\UrlAlphabet}
\g@addto@macro{\UrlBreaks}{\UrlDigits}
\makeatother

\begin{document}

%
\title{UAV Base Station Trajectory Optimization Based on Reinforcement Learning  in Post-disaster Search and Rescue Operations}

%
\author{Shiye~Zhao,~\IEEEmembership{Student Member,~IEEE,}
        Kaoru~Ota,~\IEEEmembership{Member,~IEEE,}
        Mianxiong~Dong,~\IEEEmembership{Member,~IEEE}
}


\twocolumn[
\begin{@twocolumnfalse}
  
\maketitle

\begin{abstract}
Because of disaster, terrestrial base stations (TBS) would be partly crashed. Some user equipments (UE) would be unserved. Deploying unmanned aerial vehicles (UAV) as aerial base stations is a method to cover UEs quickly. But existing methods solely refer to the coverage of UAVs. In those scenarios, they focus on the deployment of UAVs in the post-disaster area where all TBSs do not work any longer. There is limited research about the combination of available TBSs and UAVs. We propose the method to deploy UAVs cooperating with available TBSs as aerial base stations. And improve the coverage by reinforcement learning. Besides, in the experiments, we cluster UEs with balanced iterative reducing and clustering using hierarchies (BIRCH) at first. Finally, achieve base stations' better coverage to UEs through Q-learning.
\end{abstract}


\end{@twocolumnfalse}]

{
  \renewcommand{\thefootnote}{}%
  \footnotetext[1]
  {Shiye Zhao, Kaoru Ota and Mianxiong Dong are with the Department of Sciences and Informatics, Muroran Institute of Technology, Muroran, Japan (e-mail: 20043038@mmm.muroran-it.ac.jp; ota@csse.muroran-it.ac.jp; mx.dong@csse.muroran-it.ac.jp).}
  
}
 
%
\IEEEpeerreviewmaketitle

\section{Introduction}
\label{sec:I}
%
%
%
%
\IEEEPARstart{W}{ide-scale} natural disasters such as floods, tsunamis, and earthquakes occur in various places and yield destructive consequences. During sudden disasters, victims often need to contact the outside for aid as soon as possible. However, like hurricanes, flooding, or earthquakes, ground communication networks may be partially or entirely broken down\citep{xu2020big}. And the existence of aftershocks is a potential threat resulting in that the existing wireless communication networks that could not be fully relied on \citep{9136586}. Emergency communication should be deployed promptly. In the post-disaster scenario, ground transportation is usually obstructed. Emergency communication vehicles can not arrive at the scene to provide emergency communication services. The ground-based communication method is challenging to maintain reliable communication services for people in the post-disaster area. The method should be agilely implemented due to complex post-disaster situations\citep{merwaday2016stochastic}. In other words, we need reliable and agile wireless communication to act as an emergency network in any complex conditions after a disaster for post-disaster rescue and emergency communication\citep{xu2017fast}. Deploying UAVs as aerial base stations could deal with it.

\par
UAVs were primarily used in military missions, but with the development of electronic technology, UAVs were rapidly developed in reconnaissance due to their flexibility \citep{fotouhi2019survey}. Then the development of information technology broadens the range of UAV applications because they could replace humans to complete aerial operations such as logistics, agriculture, rescue, mapping, etc \citep{shakhatreh2019unmanned}. Meanwhile, greater bandwidth, lower latency, and higher density of connected mobile networks also provide more opportunities for UAVs. For instance, UAVs could act as aerial base stations to provide wireless communication to terrestrial user equipments (UEs) in the desired area \citep{mozaffari2019tutorial}. Because of the agility of UAV, UAV base stations could provide flexible communication networks at low cost reliably and efficiently \citep{li2018uav}. However, the limitation of the battery of UAVs and the time-sensitivity of post-disaster scenarios require an accurate and prompt deployment of UAVs. The issue of placement optimization should be solved \citep{lyu2016placement}. In other words, how to deploy UAVs to cover more in less time is what we focus on. Besides, the deployment should consider the TBSs partially breakdown scenario to reduce the extra cost of UAVs because some TBSs would still work after the disaster. In other words, how to deploy UAVs to cover more UEs in less time cooperating with existing TBSs is what we focus on. 
\par
\begin{figure}
    \centering
    \subfigure[Environment of gridworld]{
    \includegraphics[width=1.55in]{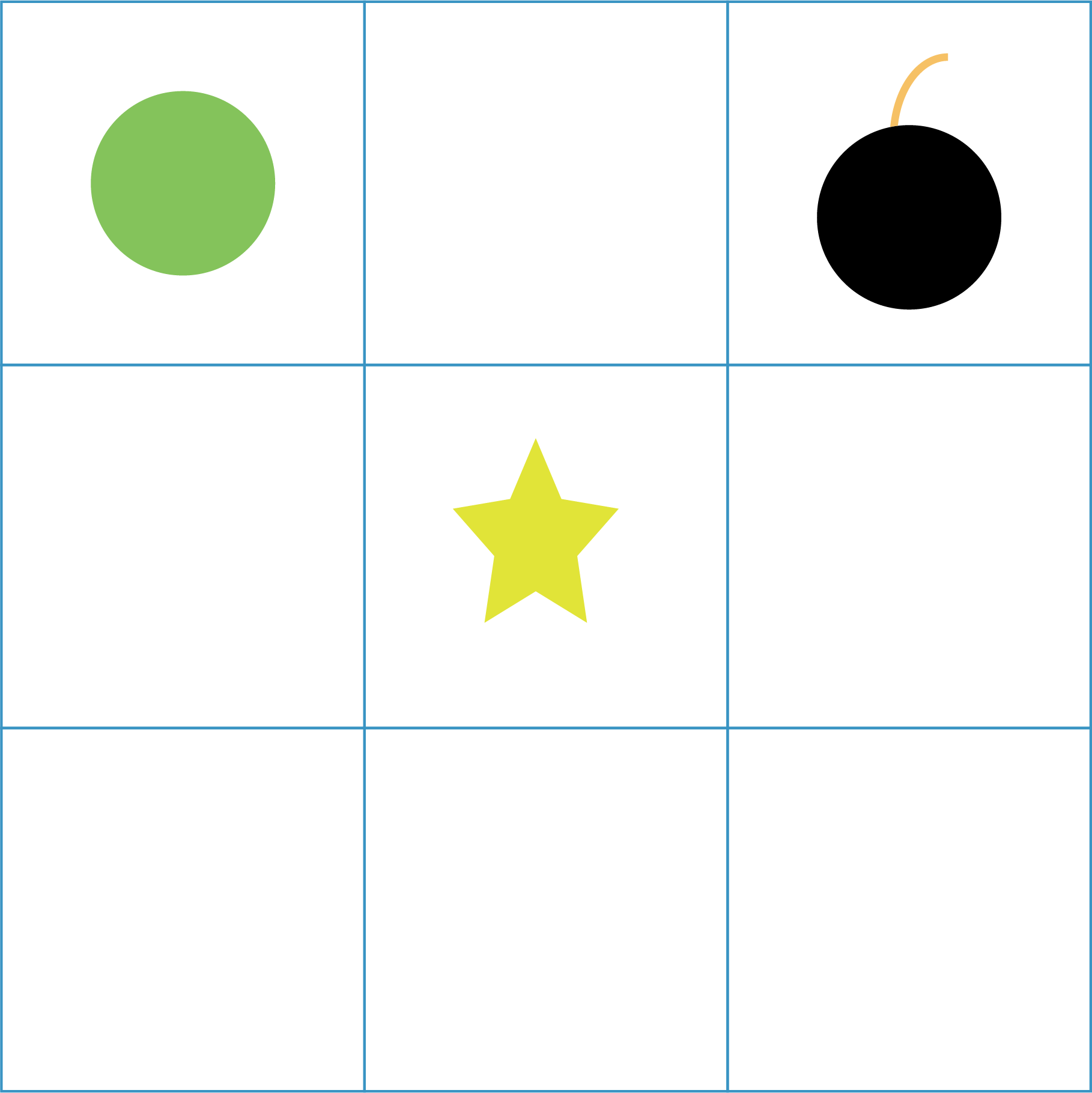}
    }
    \quad
    \subfigure[Environment of UAVs]{
    \includegraphics[width=1.55in]{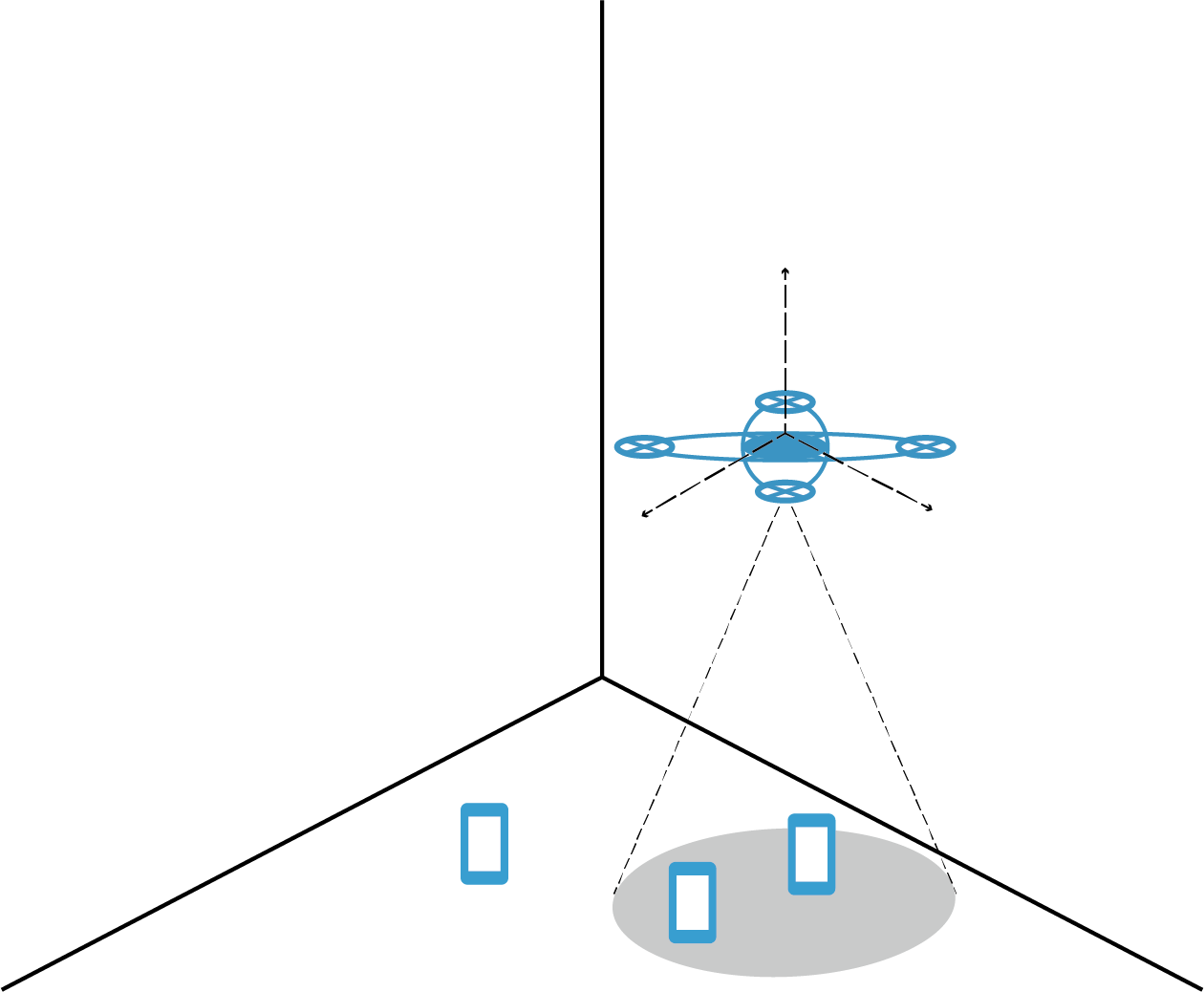}
    }    
    \caption{Environment between gridworld and UAVs}
    \label{fig:1}
\end{figure}

We propose a method to deploy UAVs as aerial base stations cooperating with these existing TBSs to serve all UEs in the post-disaster area based on Q-learning. In the post-disaster scenario, some TBSs would not be crashed down. These existing uncrashed TBSs could still continue to work to provide wireless communication services. Therefore, we could definite the connection states of UEs to classify these unserved UEs before UAVs deployment. Then deploy UAVs for UEs in the area of compromised TBSs, cooperating with existing TBSs, to build an emergency communication system to serve all UEs. In Q-learning's typical application, there is a grid world game like (a) in Figure~\ref{fig:1} in which the green circle is the player, and the player would win the game when meeting the yellow star; on the contrary, the player would lose the game when meeting the black bomb. UAVs' deployment and movement could be considered as such a three-dimension grid world game like (b) in Figure~\ref{fig:1}.  A UAV act as a player to provide emergency communication service for UEs. The player would win the game when all UEs are served within the required time and lose the game when all UEs are not covered within the required time. Meanwhile, to accelerate the steps of Q-learning, we implement the clustering algorithm with balanced iterative reducing and clustering using hierarchies (BIRCH) before Q-learning. After clustering, for each UAV, the number of UEs that need wireless communication service is decreased, then UAV's work scope is reduced, promoting UAVs to find the optimized position to cover UEs in Q-learning. Following is the contribution of this paper:

\begin{itemize}
    \item
    Consider the cooperation of TBSs and aerial base stations. There is little literature about drones' deployment in coexistence with terrestrial base stations. Existing TBSs would still provide wireless communication services to UEs. The deployment of UAVs considering existing TBSs would reduce the cost and improve efficiency. 
    \item
    Accelerate the training of Q-learning by pre-processing of BIRCH. Due to the implementation of BIRCH, the original area covered by uncrashed TBSs would be divided into smaller areas according to the coverage of the aerial base station. UAVs will find their optimal location more efficiently, accelerating the training of Q-learning.
    \item
    Deploy UAVs dynamically. UAVs are deployed based on BIRCH clustering results, and BIRCH does not require pre-input of the number of clusters. Therefore, the number of assigned UAVs would be variable with the requirement of the task. More UAVs will be assigned if the aerial base station coverage is limited or on a larger scale of maps. Otherwise, fewer UAVs would be deployed. 
\end{itemize}

\par
The rest of the paper is organized as follows. In Section \uppercase\expandafter{\romannumeral2}, we introduce the related works of UAVs' deployment. In Section \uppercase\expandafter{\romannumeral3}, we present the model of UAVs cooperating with existing TBSs to cover all UEs. Section \uppercase\expandafter{\romannumeral4}, investigates the clustering of UEs and the reinforcement learning for UAVs deployment in cellular networks. Section \uppercase\expandafter{\romannumeral5} presents the simulation result of UAVs' deployment and movement. Finally, the conclusion is presented in Section \uppercase\expandafter{\romannumeral6}.

\section{Related Works}
For industry, Amazon Prime Air\citep{amazonPrimeAir} and Project Wing\citep{projectwing} of Alphabet are outstanding cases for deploying cellular-connected UAVs in delivery service. EE and Nokia used a UAV to carry a tiny base station to provide wireless communication\citep{nokiaEE}. Google's Loon Project\citep{googleLoon}  tested to deploy UAVs for wireless communication; meanwhile, Qualcomm and AT\&T cooperated in deploying UAVs in the fifth-generation wireless network\citep{qualcommAt}. 
The literature about drone base stations has been widely developed. In \citep{feng2006wlcp2}, the authors combined building geometry and knife-edge diffraction theory to propose an A2G LoS model. \citep{al2014optimal} proposed a method to optimize a UAV's altitude for maximal coverage when the UAV acts as an aerial base station. And \citep{mozaffari2015drone} extended \citep{al2014optimal} from single UAV to two UAVs. There is also a large amount of literature about optimizing the deployment of UAV base stations. For instance, the \citep{bor2016new} discussed optimizing the 3D positions of UAVs with maximal users. The authors proposed a heuristic algorithm in \citep{kalantari2016number} to deploy UAVs with the minimum number covering all users. In \citep{lyu2016placement}, the authors proposed an algorithm to deploy UAVs in order, starting from the edge of the area and spiraling inward until it covers all UEs. \citep{mozaffari2016efficient} used Koebe–Andreev–Thurston theorem to deploy UAVs at the locations where their coverage area is maximum. \citep{kalantari2017backhaul} investigated user-centric and network-centric wireless backhaul and found the optimal backhaul-ware placement for them. 
\par
In \citep{zeng2017energy}, the authors proposed the theoretical model of the propulsion energy consumption with flight speed, direction, and acceleration. In \citep{alzenad2017}, the problem of UAV deployment was separated into vertical and horizontal dimensions to provide wireless communication for maximum UEs with minimum power. \citep{li2018deployment} used the concept of sweeping and searching user clusters to work out deploying UAVs with limited numbers to cover as many UEs as possible if there is no apriori information of user distribution. In \citep{liu2019reinforcement}, the method is put forward that reinforcement learning is used in the deployment of UAVs, obtaining the maximum quality of experience (QoE) by observing the 3D coordinates of the UAVs. \citep{hoseini2020trajectory} used particular UAVs with Q-learning to deal with UAVs' mission continuity restriction due to limited batteries. 
\par
However, these methods mentioned above solely focus on the scenario of that UAVs act as aerial base stations, and TBSs are fully broken down. As we mentioned before, after the disaster, some TBSs may still work normally. These available TBS would still provide service for UEs, and the deployment of UAVs that do not consider the existence of unbroken TBSs would cause extra cost. In conclusion, we should deploy UAVs as aerial base stations based on the post-disaster area to cooperate with existing TBSs to cover all UEs in the post-disaster area.

\section{System model}
\label{sm}

\begin{figure*}
    \centering
    \includegraphics[width=140mm]{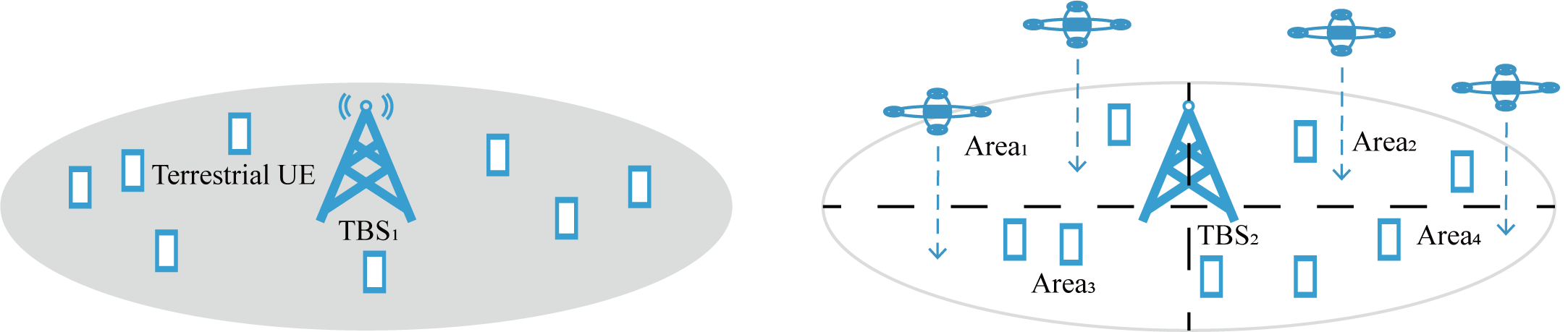}
    \caption{Post-disaster scenario}
    \label{fig:2}
\end{figure*}

The deployment of UAV for the post-disaster emergency communication is shown in Figure ~\ref{fig:2}. Before the disaster, UEs were served by TBSs like TBS$_1$. Once the TBS was crashed because of the disaster as TBS$_2$, they could not continue to serve UEs like UEs in the white circle area in Figure~\ref{fig:2}. Meanwhile, UEs served by TBS$_1$ that could continue to work would still be connected, and we do not need to deploy UAVs for them. Due to the limited coverage of UAVs, it is impossible to deploy a UAV to cover the area for an unworking TBS. Therefore, we need to cluster unconnected UEs and divide the original area covered by TBS$_2$ as Area$_{1}$, Area$_{2}$, Area$_{3}$ and Area$_{4}$ in Figure~\ref{fig:2}. Finally, deploy UAVs in these divided areas and adjust the position of UAV for better performance to connect all UEs.

\par
\begin{figure}
    \centering
    \includegraphics[width=3.5in]{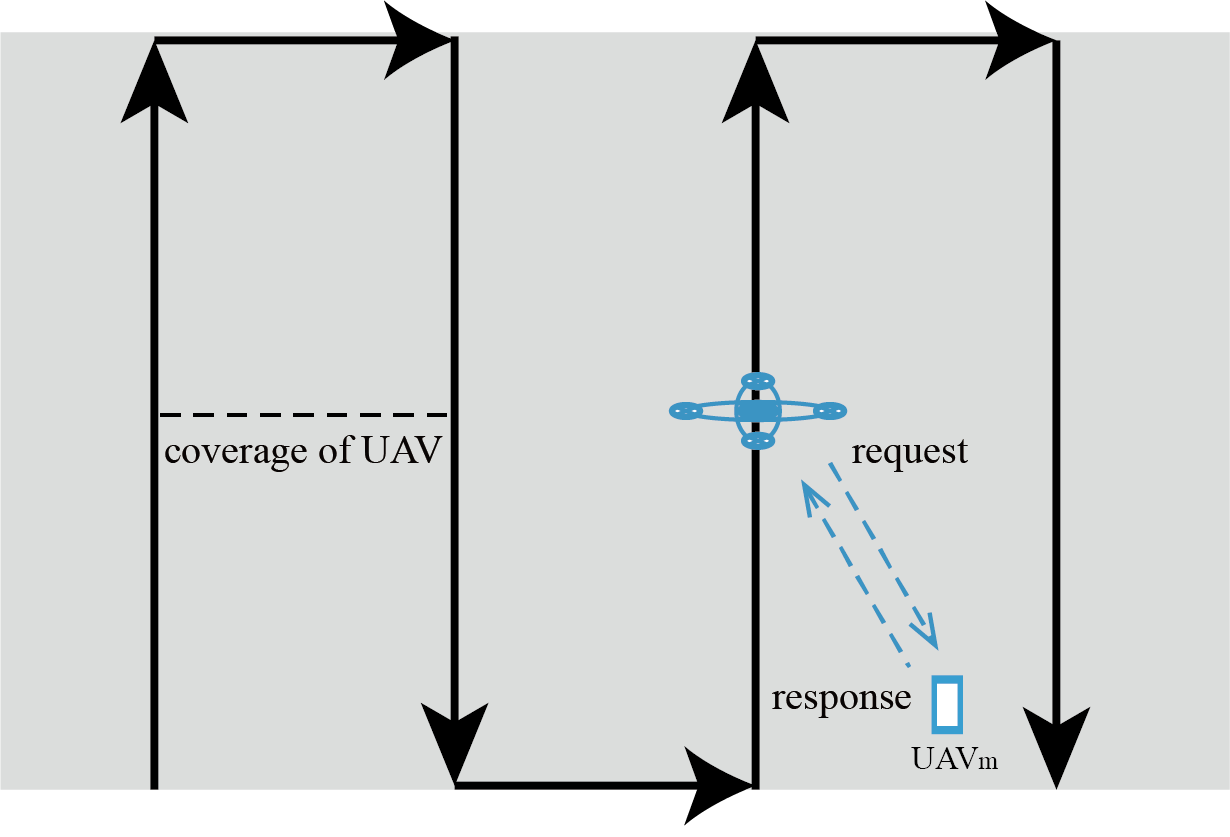}
    \caption{Mechanism of zigzag sweeping}
    \label{fig:3}
\end{figure}

To cluster these unserved UEs, we need to distinguish UEs' states whether they require a wireless emergency network or not as \(cs_{bm} \in \{0, 1\}, 0<b< N+J,0<m<M\) where b and m is the index of the \(b_{th}\) base station and the \(m_{th}\) UE,respectively, M is the amount of UE. Meanwhile, \(\sum_{b}cs_{bm} \in \{0, 1\}\). UE$_m$ would solely select one base station's service. When there is no wireless connection between transmitter$_b$ and UE$_m$, \(cs_{bm} = 0\), otherwise, \(cs_{bm} = 1\).Then there is the connection state for UE$_{m}$ and TBSs:
\begin{equation}
cs_{m} = \sum_{b=N+j}^{b=N+J} cs_{bm} 
\end{equation}
When UE$_{m}$ is unconnected with TBSs, \(cs_{m} = 0\) and \(cs_{m}=1\) if UE$_{m}$ is connected with a TBS. As we mentioned before, some TBSs could still serve UEs after the disaster, and \(cs_{m} = 1\) for these UEs. UAVs could be deployed to serve UEs whose \(cs_{m} = 0\). Besides, we need the position of UEs when we cluster them. A mechanism to scan the environment and obtain the states and information of UEs should be implemented here. To cover the whole map, we deploy UAV to scan the area with the zigzag sweep as Figure~\ref{fig:3} with the interval based on coverage of UAVs. During the zigzag sweep, UAVs would broadcast request messages. Then UEs with \(cs_{m} = 0\) would respond to the request message. The response message would contain information of UEs \(U=\{U_1, U_2, ..., U_m, ..., U_M\}, U_{m} = \{loc_{m}, cs_{m}\}\) where \(loc_{m}\) is the location of UE$_{m}$, and the center would cluster them based on their information as \(Cluster = \{clus_{1}, clus_{2}, ...,  clus_{n}, ...,clus_{N}\}, clus_{i}\cap clus_{j} = \emptyset\) if \(i \neq j\). 

\par
\begin{figure}
    \centering
    \includegraphics[width=3.5in]{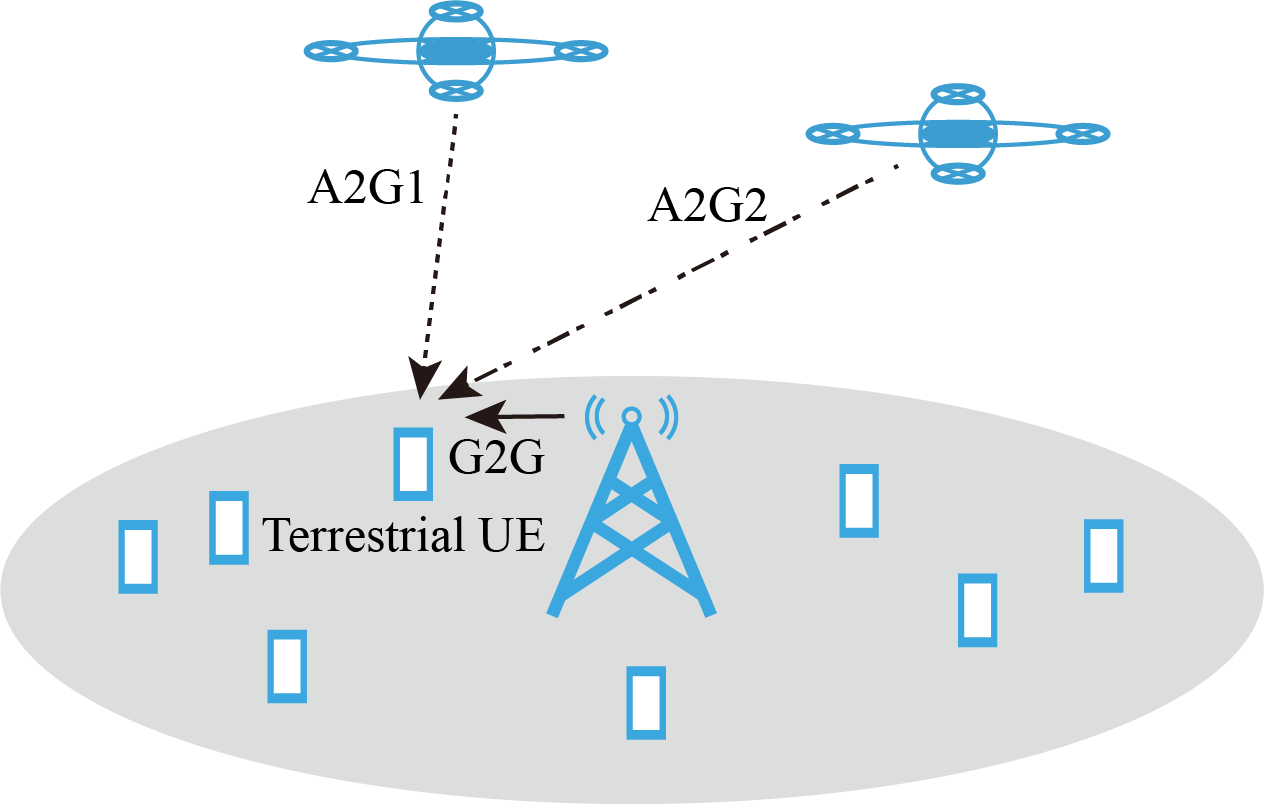}
    \caption{Scenario of A2G and G2G model}
    \label{fig:4}
\end{figure}

After clustering, UAVs would adjust locations to provide better service for UEs. The transmission of UAVs-assisted cellular networks is shown as Figure~\ref{fig:4} in which we deploy muti-UAVs cooperating as aerial base stations with existing TBSs to cover all victims in the post-disaster area. For the Terrestrial UE in Figure~\ref{fig:4}, the received signals consist of 2 parts, signals from aerial base stations to the ground UEs (A2G) expressed as A2G$_1$ and A2G$_2$ respectively in Figure~\ref{fig:4}, and the TBS to ground UEs (G2G) channel expressed as G2G in the Figure~\ref{fig:4}. After receiving signals from different base stations and comparing them, the user chooses to build a wireless communication link with the base station to provide the best service. In other words, if A2G$_1$ perform best for Terrestrial UE, both of A2G$_2$ and G2G are interference signal to Terrestrial UE. Similarly, if A2G$_2$ performs best, A2G$_1$ and G2G are interference, and if G2G performs best,  A2Gs are interference for Terrestrial UE.

For better service, the UEs will compare the performance of base stations. The closer UAV would provide better service because of less loss and stronger signal. What we need first is calculating the distance from UEs to base stations. We formulate \(BS=\{UAV_1, ..., UAV_n, ...,UAV_N,  TBS_1, ..., TBS_j, ..., TBS_J\}\) to denote base stations, including UAVs and TBSs, where \(N\) and \(J\) are the numbers of UAV and TBS, respectively. There is a distance of each UE to a base station that is formulated as  

\begin{equation}
    d_{bm} = \sqrt{(x_b - x_m)^2 + (y_b - y_m)^2 + z_b^2 } ,
\end{equation}

where \(x\), \(y\), and \(z\) are values of three-dimensional coordinate, and \(b\) denotes the index of $b_{th}$ base stations. Meanwhile, we formulate \(D_{um}=\{d_{1m},d_{2m}, ..., d_{nm}, ..., d_{Nm}\}\) and \(D_{mm}=\{d_{(N+1)m}, ..., d_{(j)m}, ...,d_{(N+J)m} \}\) to denote distances from  UAVs and TBSs.

\par
For UE$_m$ with \(cs_{m}=1\), the connection is based on the distances between TBSs and UE$_m$. We redefine the distances between TBSs and UEs as 
\begin{equation}
    dm_{jm}\! \! = \! \! \begin{cases}
        d_{(N+j)m}, \! cs_{(N+j)m} \!= \!1 \! \!\!\And\!\! \!d_{(N+j)m}\! \leq\! d_{thrbs},\\
        \infty,  others.\\
    \end{cases}
\end{equation}

For the convenience of calculation, the length is set to infinite when the user is out of the TBS's coverage. When selecting a base station, UE$_{m}$ would consider TBSs whose distance \(dm_{jm}\) meets.

\begin{equation}
    dm_{jm} = \min_{j}dm_{jm}, \qquad dm_{jm} \neq \infty
\end{equation}

\par
Considering the propagation of G2G would combine with the reflection, we choose the Two-Rays Ground Model in wireless communication to represent the propagation of the G2G channel in our model: \begin{equation}
    P_{rbjm} = 10 \log(\frac{|E_{jm}|^2}{\eta}A_{rj}).
\end{equation}
\(\eta\) is a constant denoting the impedance of free space, and \(A_{rj}\) is the effective surface of receiving antenna that could be calculated by \(A_{rj}=\frac{\lambda_{tbsj}^2}{4\pi}G_r\), where \(\lambda_{tbs}\) is wavelength of signals from terrestrial base stations, and \(G_r\) is the gain of the receiving antenna from terrestrial users. Besides, \(E_jm\) is the electric field strength formulated by 
\begin{equation}
    E = \frac{4\pi h_{j} h_{m}\sqrt{30P_{t}G_{t}}}{\lambda_{js} d_{jm}^{'2}},
\end{equation}
where \(G_{t}\) and \(P_{t}\) are the gain and transmission power of the transmitting antenna, respectively, \(h_j\) and \(h_m\) are the heights of the transmitter and receiver, and \(d'_{jm}\) is the distance between them.

\par
Because UEs served by normally working TBSs after the disaster would not re-select a UAV to connect when calculating the minimum distance to choose an aerial base station, we just consider \(D_{um}\) when unconnected UEs select base station.

\par

For UE$_m$ with \(cs_{m}=0\), UEs' initial connection state denotes damaged conditions of terrestrial base stations because users are keeping connecting to terrestrial base stations and \(cs_{m}=1\) before the disaster, once the base station is crashed \(cs_{m}=0\). Therefore, aerial base stations would build emergency wireless communication networks for UEs with \(cs_{m}=0\). Every time UAVs moving, if distances between the UAV and UEs in the area where the UAV is assigned are below the maximum range of UAVs d$_{thruav}$ as

\begin{equation}
    cs_{nm}\! \! = \! \! \begin{cases}
        1, \!  U_{m} \in clus_{n} \! \And \!du_{nm}\! \leq\! d_{thruav},\\
        0,  others.\\
    \end{cases}
\end{equation}

Similarly, we also formulate the distances between UAVs and UEs as

\begin{equation}
    du_{nm}\! \! = \! \! \begin{cases}
        d_{nm}, \qquad d_{nm}\! \leq\! d_{thruav},\\
        \infty,  others.\\
    \end{cases}
\end{equation}

The UAVs' transmission of signals in the A2G channel contains two parts: line of sight propagation (LOS) and none line of sight propagation (NLOS). UEs would receive signals containing the two parts mentioned above. And transmissions of LOS and NLOS propagation occur with different possibilities separately. The possibility of LOS link in the A2G channel is formulated as
\begin{equation}
    P_{los}(\theta_{nm}) = \frac{1}{1 + \alpha \exp(-\beta[\theta _{nm} - \alpha])} ,
\end{equation}
where \(\alpha\) and \(\beta\) are constant values determined by the environment, such as countryside or city, besides \(\theta_{nm}\) is the angle between the UAV and UE, which is computed by \(\theta _{nm} = \arcsin(y_n/du_{nm})\). Then we could calculate the possibility of the NLOS link from 
\begin{equation}
    P_{nlos}(\theta_{nm}) = 1 - P_{los}(\theta_{nm}) .
\end{equation}
In the propagation model, because the multipath shadowing caused by the multiple reflected signals is less than other components, we just focus on the path loss of LOS, NLOS, and free space propagation. The path loss between aerial base station with terrestrial UEs in the propagation model could be formulated by
\begin{equation}
    L_{nm} = 20\log(\frac{4\pi f_c du_{nm}}{c}) + P_{los} \mu _{los}+P_{nlos} \mu _{nlos},
\end{equation}
where \(c\) is the velocity of light and \(f_c\) is the carrier frequency of signals, besides \(\mu_{LOS}\) and \(\mu_{NLOS}\) are the constant value represented the attenuation in LOS and NLOS. With the path loss of aerial base station to terrestrial users we calculated above, the power between them could also be calculated by
\begin{equation}
    P_{rnm} = 10\log P_{tn} - L_{nm},
\end{equation}
where \(P_{tn}\) is the transmission power of the aerial base station. And \(P_{rnm}\) is the received power by terrestrial users.

\begin{figure*}
    \centering
    \includegraphics[width=140mm]{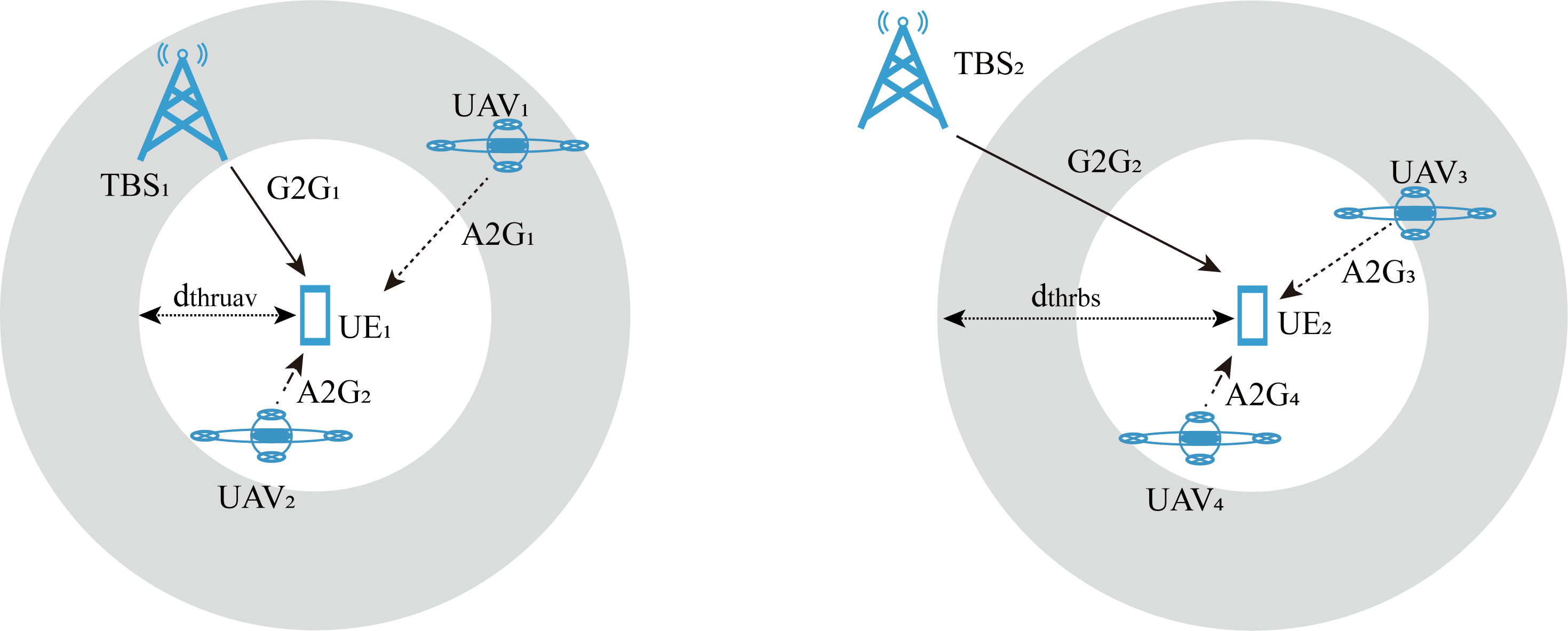}
    \caption{Interference with minimum distance and threshold}
    \label{fig:5}
\end{figure*}

\par
As Figure~\ref{fig:4} shows, UEs on the ground would always receive signals from base stations, like A2G$_1$, A2G$_2$, and G2G. We assume the method of base station selection is based on the distances mentioned above. Interference for UE$_m$ is shown as Figure~\ref{fig:5}. 

\par
In Figuire~\ref{fig:5}, the inner and external radiuses of circles d$_{thruav}$ and d$_{thrbs}$ are the maximum connection distance of UAVs and TBSs. For UE$_1$, it is in the coverage of active TBS$_1$; therefore, A2G$_1$ and A2G$_2$ from UAVs are interferences. Meanwhile, UE$_2$ is out of the coverage of TBS$_2$ but covered by UAV$_3$ and UAV$_4$. Assume that UE$_2$ is in the cluster of UAV$_4$ \(clus_{4}\), both of G2G$_2$ and A2G$_3$ are interferences for UE$_2$. The power of the effective signals is \(P_{e_1}=P_{rb11}\) and \(P_{e_2}=P_{r42}\); meanwhile, the power of the interfering signals is \(P_{i_1} = P_{r11}+P_{r21}\) and \(P_{i_2} = P_{rb22}+P_{r32}\). On account of what we formulated above, the power of connection and interference could be obtained. Then the signal to interference plus noise ratio (SINR) could be calculated by 

\begin{equation}
    SINR_m = \frac{P_{e_m}}{P_{i_m} + P_{noise}},
\end{equation}
where \(P_{noise}\) is the interference power of noise. 
\par
Meanwhile, the rate for users could be calculated by Shannon's formula, which is formulated as
\begin{equation}
    rate_{bm} = \frac{B_b}{M_b}\log_2(1+SINR_m),
\end{equation}
\(B_b\) is the bandwidth of the transmitter, and \(M_b\) is the number of receivers for the transmitter$_b$. For UE$_m$, the final rate could also be denoted by 
\begin{equation}
    rate_m=\sum_{b}r_{bm}cs_{bm},
\end{equation} 
what we force to solve is maximizing the \(M_n\) with higher \(rate_m\) as 
\begin{equation}
\begin{split}
&\max (\sum_{n}M_{n}+\sum_{j}M_{j}).\\
&s.t.\quad  \left\{\begin{array}{lc}
s_{n}\in S_{n}\\
s_{n}\neq s_{n'} \quad if \quad n \neq n' \\
t_{n}<t_{th}\end{array}\right.
\end{split}
\end{equation}
It denotes we would achieve that most victims could receive UAVs' available service within the required time $t_{th}$; meanwhile, the position of $UAV_{n}$ is in its assigned area $S_{n}$, and for different UAV their positions $s_{n}$ and $s_{n'}$ should be not same.

\section{Proposed algorithm}

\label{pa}

\begin{figure}
    \centering
    \includegraphics[width=3.5in]{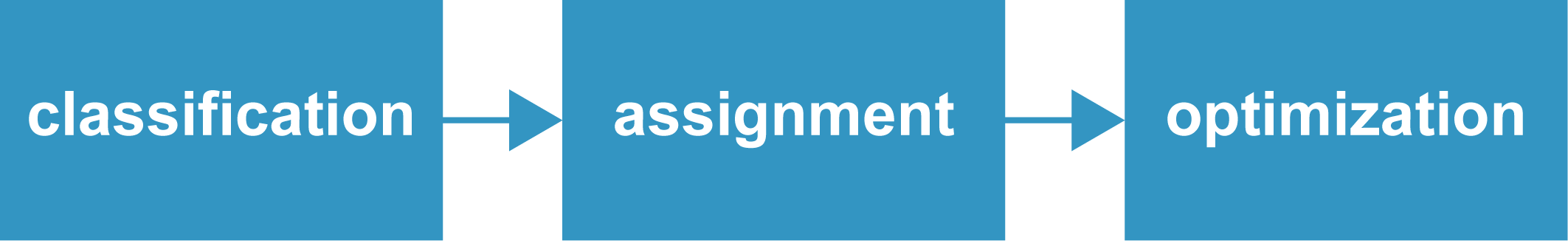}
    \caption{Steps of method}
    \label{fig:steps}
\end{figure}

To cover more UEs after the disaster, we propose the methods containing three steps to deploy aerial base stations in the scenario that TBSs are partially crashed down, which is shown in Figure~\ref{fig:steps}. As we mentioned before, we do not need to deploy UAVs for these UEs covered by available TBSs, so we should classify which TBS does not work at the beginning then pick these unserved UEs before other steps. Then we design the appropriate assignment of UAVs considering the coverage of aerial base stations. Finally, optimize trajectories of UAVs to obtain the maximum amount of UEs we talked about in Section~\ref{sm}.

\par
\begin{figure}
    \centering
    \includegraphics[width=3.5in]{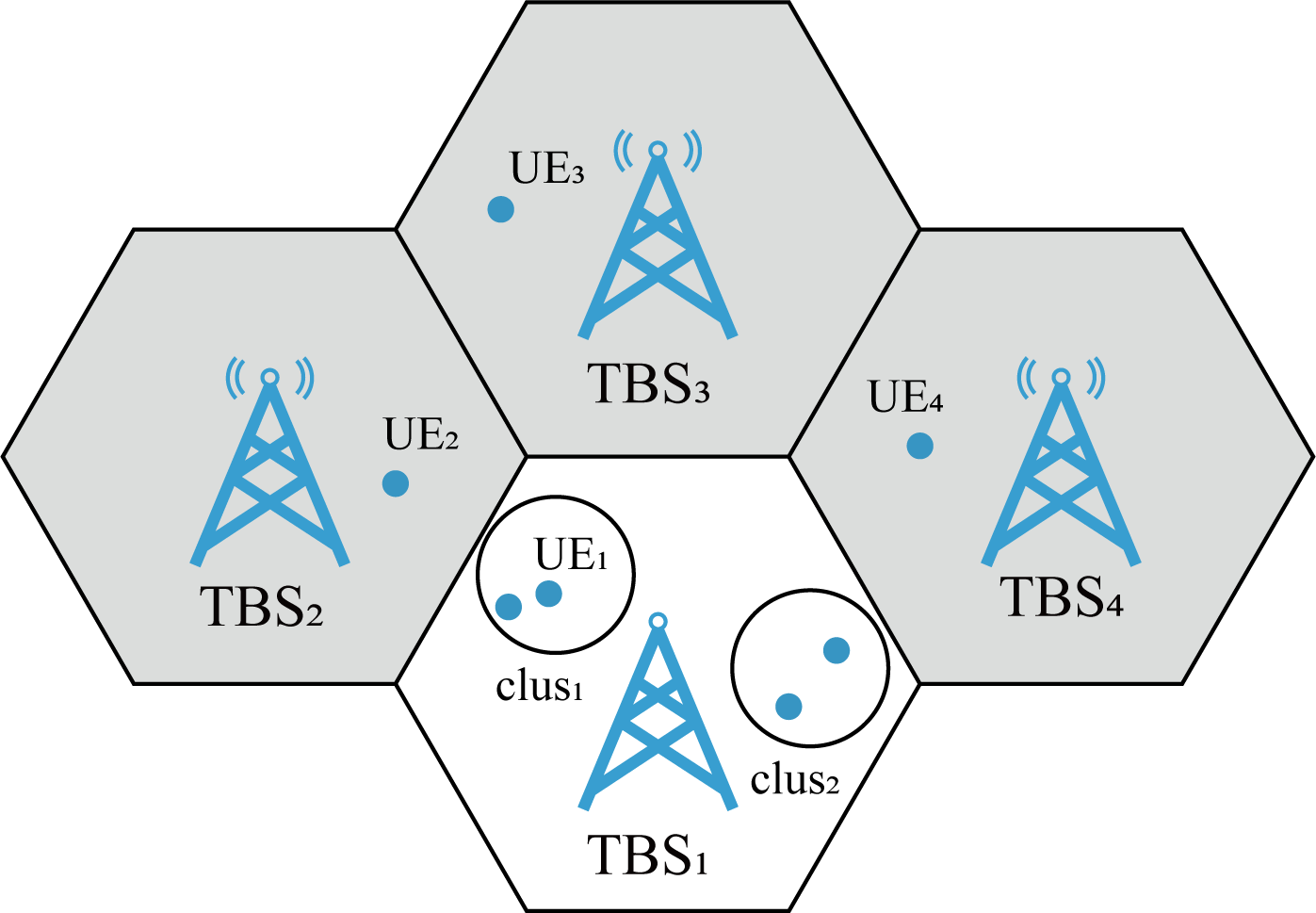}
    \caption{Division of area for UAVs}
    \label{fig:6}
\end{figure}

Before deployment of UAVs, we need to detect the area to require unconnected UEs' location first. With the location of unconnected UEs, we cluster them with the clustering algorithm as Figure~\ref{fig:6}. Before the disaster, all UEs were served by TBSs as usual. Because of the impact of the disaster, TBS$_1$ does not continue to work, UEs served by TBS$_1$ are unconnected. To serve all UEs in the post-disaster area, we need to build an available emergency wireless communication network. For connected UEs as UE$_2$, UE$_3$, UE$_4$, they are still connected with an TBS like TBS$_2$, TBS$_3$, TBS$_4$, in Figure~\ref{fig:6}. We need to deploy aerial base stations to cover the unconnected UEs as UE$_1$ in Figure~\ref{fig:6} but those connected UEs.
\par
In Section~\ref{sm}, we formulate UEs information as \(U_{m} = \{loc_{m}, cs_{m}\}\), then deploy UAVs with zigzag sweeping to obtain all UEs information. With \(U_{m}\) we obtained, we could pick up these unserved UEs whose \(cs_{m} = 0\) directly and compelete the mission of classification.
\par
Because the coverage of the UAV is smaller than the TBS, we need to cluster unconnected UEs and divide the area covered by TBS$_1$. In Figure~\ref{fig:6} UEs are clustered into two cluster: $clus_1$ and $clus_2$, where circles encircling UEs in a cluster is the range of movement for UAVs. To deploy UAVs dynamically, we cluster UEs with Balanced Iterative Reducing and Clustering Using Hierarchies (BIRCH). BIRCH cluster data based on Clustering Feature (CF) tree that is combined by nodes with CFs. A CF is a linear ternary array containing the number of samples, the sum vector of each characteristic dimension of each sample, and the sum of squares of them, in which the CF of a parent node equals the sum of all CFs in child nodes. The process of BIRCH is building a CF tree with all data. When adding new data, searching the closest CF in leaf nodes, if the data is in the coverage of CF under radius threshold, add the data in the closest CF leaf node, else the data would be added in a new CF in the leaf node. When adding the data into the CF tree, if the amount of the CFs for non-leaf nodes is more than the default maximum amount of CF for non-leaf nodes, divide the original node into two nodes and select two CFs with the farthest distance in the original node as the first CF in the new node, then add other nodes based on the distance in two leaves. After adding the data into the new nodes, update the CF tree. Similarly, if the amount of CF of leaf nodes is more than the maximum number of CFs for leaf nodes, divide it into two new nodes and assign CFs for them. Then we could obtain the clusters in each leaf node.
\par
After clustering, divide the area based on the minimal enclosing circle of each cluster to reduce the area where UAVs work as \(clus_1\) and \(clus_2\) in Figure~\ref{fig:6}. 
\par

\begin{algorithm}
\caption{Pre-prosess of UAVs' deployment}
\label{alg:pre}
\begin{algorithmic}
\REQUIRE {locations of UEs and states of TBSs}
\STATE{move UAV to cover the post-disaster area based on zigzag}
\STATE{store informations of $UE_{m}$ whose $cs_{m}=0$}
\ENSURE{for $UE_{m}$ whose $cs_{m}=0$}
\STATE{find the closest node and Cluster Feature $CF_i$}
\IF{$UE_{m}$ is in the coverage of $CF_i$}
\STATE{add $UE_{m}$ into $CF_i$}
\ELSE
\STATE{create $CF_{j}$ and add $UE_m$ into $CF_{j}$}
\IF{the amount CFs of the current node is more than the threshold}
\STATE{split the node into two nodes}
\STATE{assign CFs of the original node into new nodes}
\ENDIF
\ENDIF
\STATE{update CFs in the route of current node}
\STATE{obtian clusters in leaves as $Cluster=\{clus_{1},..., clus_{n},...,clus_{N}\}$}
\STATE{find the minimal enclosing circle for each cluster}
\STATE{index \(clus\) and assign a UAV}
\end{algorithmic}
\end{algorithm}

\par
In \citep{watkins1992q}, the authors proposed a kind of reinforcement learning algorithm based on value named Q-learning. The agent could get a reward in the sate \(s(t)\) and choose an action \(a\) by every step in Q-learning. The agent builds a Q-Table to store the state, action, and Q value that is denoted as \(q(s_t,a)\), finally choosing the action to reach the maximum reward based on Q-Table. Here, we assume that the problem of UAVs' deployment is a Markov Decision Process (MDP), the future state \(s_{t+1}\) is only decided by the current state \(s_t\), then the problem could be dealt with Q-Learning.

\par
The state in the model is:  
\begin{equation}
    S(t)=\{s_1(t), s_2(t),..., s_n(t),..., s_N(t)\},
\end{equation}
where \(s_n(t)\) denotes the state of agent$_n$ and UAVs are agents here. The amount of agent $N$ would be related to the number of clusters in BIRCH. For every UAV, the state includes the Cartesian coordinates of UAVs in the scene, which are formulated by
\begin{equation}
 	s_n(t) = \{x_n(t), y_n(t), z_n(t)\}.
\end{equation}
The distances between users to UAVs could be calculated every time the UAVs move, and rewards could be calculated meanwhile.

\par
When getting information of UAV$_n$' locations, agent$_n$ could choose an action in the action space based on \(s_n(t)\). The action space of UAVs could be denoted as 
\begin{equation}
    A(t) = \{a_1(t), a_2(t),..., a_n(t),..., a_N(t)\}
\end{equation}
where \(a_n(t)\) is the action of UAV$_n$, and it is defined as \(a_n(t)\in\{0,1,2,3,4,5,6\}\) representing moving front, back, left, right, higher, lower, or hovering, respectively. After the agent choose an action from \(a_n(t)\in\{0,1,2,3,4,5,6\}\), states of agents become \(S(t+1)\), and repeat steps of choosing action at different states until it meets the restrict of steps that denote UAVs run out of the battery.

\par
As we discussed in Section~\ref{sm}, \(M_n\) and \(rate_n\) are the number of connected users and users' rates provided by UAV$_n$ respectively. \(M_{n}\) would increase while the UE's rate is over the threshold of rate \(rate_{th}\) we set. With the threshold of rate \(rates_{th}\), we could broaden aerial base stations' coverage, providing a reliable wireless communication service. The reward should be settled to increase users' amount and rates. Then we could denote the reward function by
\begin{equation}
    r_n(t) = \begin{cases}
        0, \quad M_n(t) \geq M_{total_n}, \\
        \frac{M_{n}(t)}{M_{total_n}} - 1, \quad others,\\
    \end{cases}
\end{equation}
where \(M_{total_n}\) is the total number of UEs served by UAV$_n$. UAVs would explore to cover more UEs faster with a reliable wireless communication service while training.

\begin{algorithm}
\caption{Movement of UAVs based on Q-learning}
\label{alg:SA}
\begin{algorithmic}
\REQUIRE {the environment and Q-table}
\STATE{deploy agent into areas divided by pre-processing}
\ENSURE{for each episode}
\STATE{initialize $s$ for each UAV}
\FOR{ steps in the episode}
\IF{$\epsilon_n$ is greater than $\epsilon_{nmin}$}
\STATE{$a_n \leftarrow  \arg\max Q(s_n, a_n)$}
\ELSE
\STATE{generate $a_n$ in action space randomly}
\ENDIF
\STATE{observe $r$ and $s'$}
\IF{$s'_n$ equals to other uavs' $s'$}
\STATE{$s'_n \leftarrow   s_n$}
\ENDIF
\STATE{observe $r$ and $s'$}
\STATE{$Q(s, a) \leftarrow Q(s, a) + \alpha \left\{r + \gamma \max [Q(s', a')-Q(s, a)] \right\} $}
\STATE{$s \leftarrow s'$}
\STATE{$\epsilon_n \leftarrow \epsilon_n de$}
\ENDFOR
\end{algorithmic}

\end{algorithm}

\par
After the clustering and workplace dividing, the interference from other agents would be reduced, and UAVs could be deployed efficiently. The agents here are UAVs and independent from each other. Every time the agents are observing the environment, they would act respectively based on their Q-tables. Choosing action by Q-tables here, we design it with epsilon greedy policy as
\begin{equation}
    \pi_n(s_n|a_n)\!=\!\begin{cases}
        1\!-\!\epsilon\!+\!\frac{\epsilon_n}{k}, \ a_n = {\arg\max}Q(s_n,a_n),  \\
        \frac{\epsilon_n}{k}, \quad others.\\
    \end{cases}
\end{equation}
where \(\pi_n(s_n|a_n)\) is the possibility of an action in \(s_n\), and \(k\) is the amount of actions. Because the agent does not have enough information about the environment, we want it to be explored randomly initially. 

\par
Therefore, we use a decay parameter as \(\epsilon_n = de * \epsilon_n\), where \(de\) is a constant set by us, \(\epsilon\) would decay every episode. Besides, we would settle a constant, the threshold of decay parameter, \(\epsilon_{nmin}\). Once \(\epsilon_n\) decline below \(\epsilon_{nmin}\), the decay parameter would set as \(\epsilon_{nmin}\) as
\begin{equation}
    \epsilon_{n}=\begin{cases}
        \epsilon_{n}, \quad \epsilon_{n} \geq \epsilon_{nmin},  \\
        \epsilon_{nmin}, \quad \epsilon_{n} < \epsilon_{nmin}.\\
    \end{cases}
\end{equation}
Then, the agent would explore randomly in an unfamiliar environment then act as epsilon greedy. All users would connect to the transmission device based on the protocol mentioned above. The reward would update by the number of users and rates. Furthermore, we add collision detection here. If a UAV's next state is the same as $s'_n$, it would be determined as a collision in the next moment. Then the latter one's action would change to stay in this moment, and the prior one continues to act. After collision detection, agents would update the Q-table and state based on the final reward $r$ and state $s'$. Due to agents working independently, as we mentioned above, the reward is respective. We formulate the performance of the whole system as
\begin{equation}
    r_{sys}=\sum_{n}r_{n}.
\end{equation}
With the system's reward \(r_{sys}\), we could evaluate the system's performance and make all UEs served.

\section{Simulation result}
We set the simulation environment in a 5200*5200*5200 grid world. There are 4 TBSs with state \(\{0, 1, 1, 0\}\) in this simulation. The coverage of TBSs and UAVs is 2400, 1200, respectively. Table~\ref{tb1} shows the parameters of the environment and base stations. We assume the center frequency of the TBSs and UAVs is \(f_{c}=2Ghz\), and transmitter power is \(P_{t}=4kW\). TBSs' transmitter gain and UEs' receive gain are \(G_{t}=G_{r}=1\). All channels' bandwidth is \(B_{n} = 0.18Mhz\). Meanwhile, we assume that all TBSs' height is \(h_{t}=100m\), and UAVs' initial coordinates are randomly in the assigned area. We set the workplace of UAVs in urban; therefore, the constant values \(\alpha=1\) and \(\beta=1\). UAVs' attenuation factors of the LOS link and NLOS link are \(\mu_{los}=3\), \(\mu_{nlos}=23\), respectively. Besides, noise's power spectral density used to calculate noise's power is \(-174dBm/HZ\).

\par
\begin{table}
    \centering
    \caption{Parameters of Environment}
    \label{tb1}
    \begin{tabular}{c|c}
    \hline
        Parameter & Value\\
    \hline
        $f_c$& 2$Mhz$  \\
    \hline
        $P_t$& 4$kW$  \\
    \hline
        $G_t$& 1  \\
    \hline
        $G_r$& 1  \\
    \hline
        $B_n$& 0.18$Mhz$\\
    \hline
        $N_0$& -174$dBm/HZ$\\
    \hline
        $h_t$&100$m$\\
    \hline
        $\beta$&1\\
    \hline
        $\alpha$&1\\
    \hline
        $\mu_{los}$&3\\
    \hline
        $\mu_{nlos}$&23\\
    \hline
        $n$&5\\
    \hline
    \end{tabular}
\end{table}

\par
Meanwhile, we set the parameter of Q-learning as Table~\ref{tb2}. The learning rate is \(lr = 0.5\), and reward decay is \(\gamma = 0.9\). Besides, the initial parameter of the epsilon-greedy algorithm is \(\epsilon = 0.9\), and the decay parameter is \(de = 0.999\). The threshold of $\epsilon$ is \(\epsilon_{min} = 0.01\). Then set \(ep=10000\), and simulate 10000 episodes.
\begin{table}
    \centering
    \caption{Paramaters of Q-learning}
    \label{tb2}
    \begin{tabular}{c|c}
    \hline
        Parameter & Value\\
    \hline
        $\gamma$& 0.9  \\
    \hline
        $\epsilon$& 0.9  \\
    \hline
        $de$& 0.999  \\
    \hline
        $\epsilon_{min}$& 0.01  \\
    \hline
        $lr$& 0.5\\
    \hline
    \end{tabular}
\end{table}

\par
\begin{figure}
    \centering
    \includegraphics[width=3.5in]{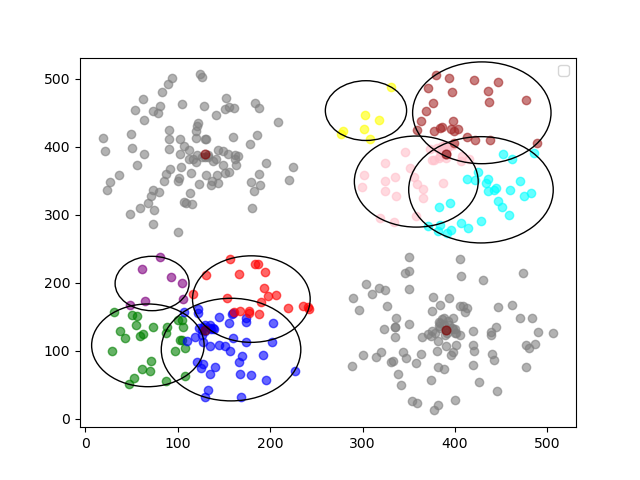}
    \caption{User clusters with Birch}
    \label{fig:7}
\end{figure}

Figure~\ref{fig:7} plots the clusters of UEs when the TBS's state is {0,1,1,0}. In the simulation, there are 400 UEs. Because base stations cover all people before the disaster, users distribute around base stations in base stations' coverage area. We assume that users distribute around TBSs in Poisson Point Process (PPP) geographically. Then we combine UEs' distribution of every base station, as Figure~\ref{fig:7} shown. The black points in the graph represent UEs served by the TBSs after the disaster, and four maroon spots are TBSs. Because the state of TBSs is \(\{0,1,1,0\}\) as we mentioned above, users who need emergence wireless service provided by UAVs are mainly distributed in coverage of TBS$_1$ and TBS$_4$. That is to say, UAVs' workplace is two 260*260*260 grid space, and TBSs serve users in the rest part. Since the scale in the environment is 1:20m, the UAVs would work in two 5200m*5200m areas. As mentioned above, the transmission distance of the UAV is 1500m. To ensure that the UAVs cover the whole assigned area, we cluster UEs with the coverage of the UAV with BIRCH, and unconnected UEs are divided into eight clusters. Circles filled by colorful points in the figure are the area for a UAV to serve unconnected UEs. And these colorful points are clusters after BIRCH. The amount of clusters is eight; therefore, we would deploy eight UAVs to support TBSs to cover all UEs in this post-disaster area.

\par

\par

\begin{figure} 
	\centering  
	\vspace{-0.35cm} 
	\subfigtopskip=2pt 
	\subfigbottomskip=2pt 
	\subfigcapskip=-5pt 
	\subfigure[agent0]{
		\label{reward0}
		\includegraphics[width=0.875in]{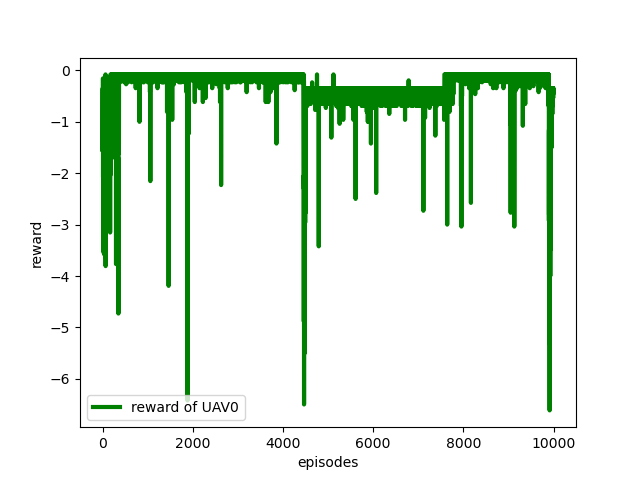}}
	\subfigure[agent1]{
		\label{reward1}
		\includegraphics[width=0.875in]{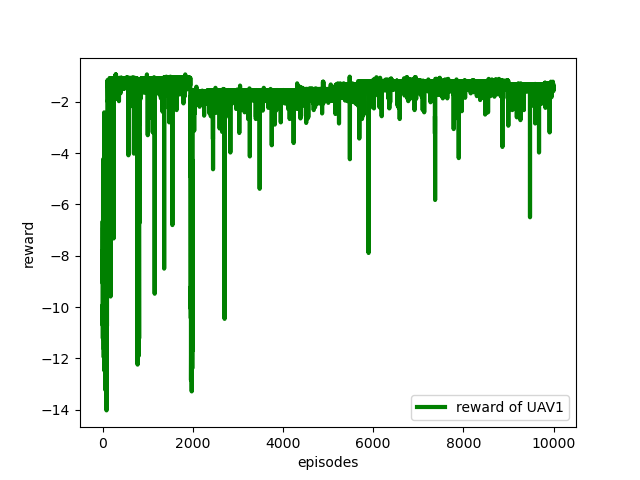}}
	\subfigure[agent2]{
		\label{reward2}
		\includegraphics[width=0.875in]{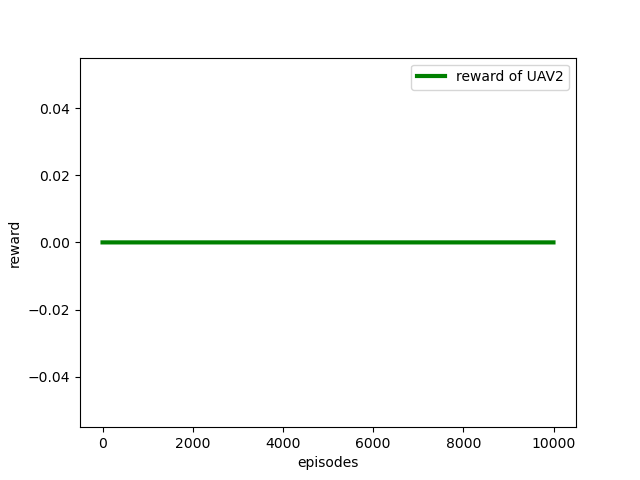}}
	\subfigure[agent3]{
		\label{reward3}
		\includegraphics[width=0.875in]{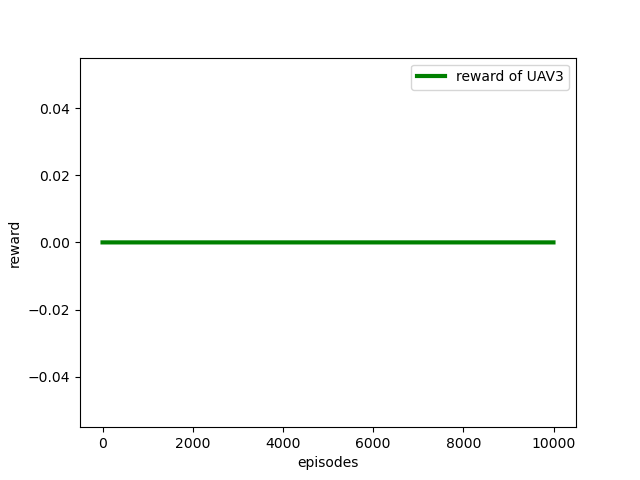}}
	\subfigure[agent4]{
		\label{reward4}
		\includegraphics[width=0.875in]{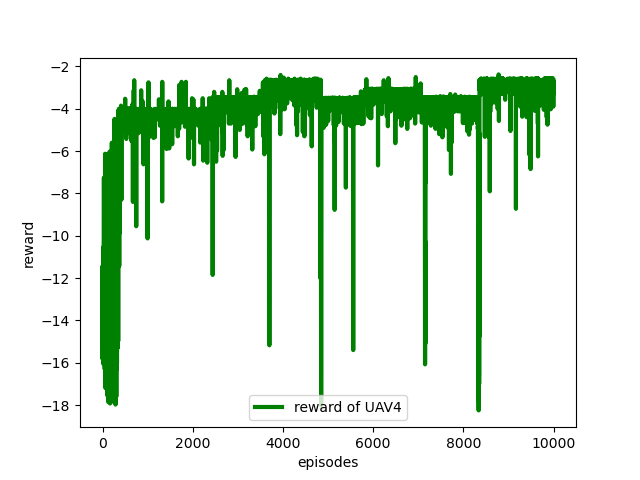}}
	\subfigure[agent5]{
		\label{reward5}
		\includegraphics[width=0.875in]{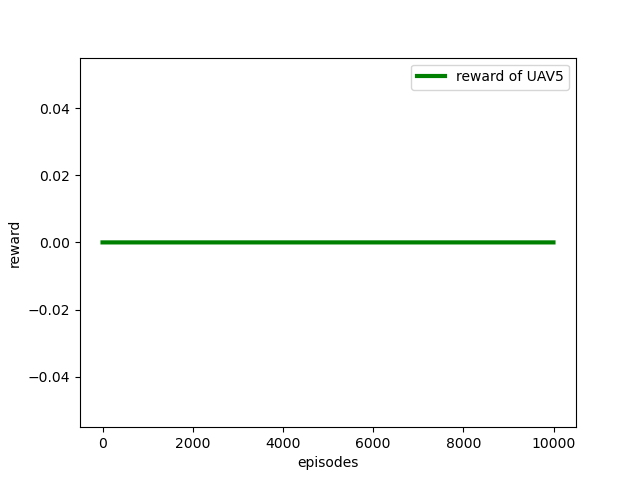}}
	\subfigure[agent6]{
		\label{reward6}
		\includegraphics[width=0.875in]{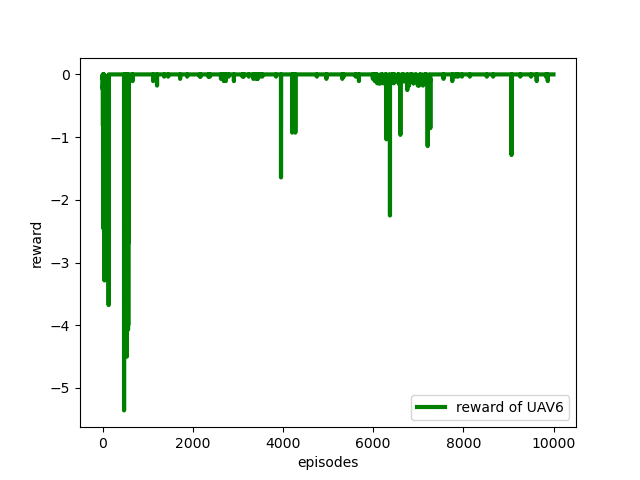}}
	\subfigure[agent7]{
		\label{reward7}
		\includegraphics[width=0.875in]{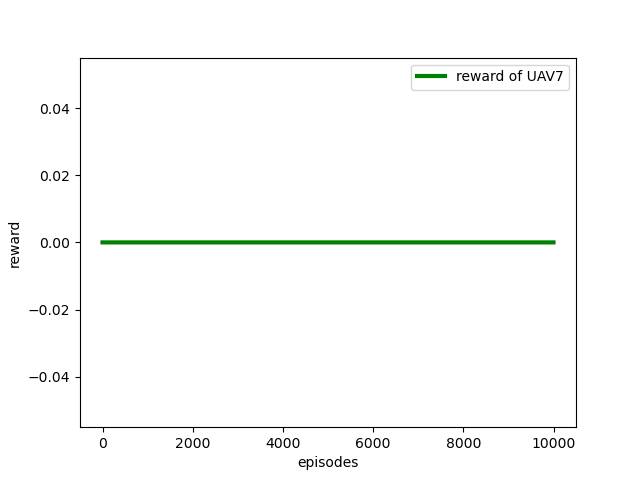}}
	\caption{Reward for each agent}
	\label{fig:8}
\end{figure}

Fig~\ref{fig:8} shows the reward for each agent. Fig~\ref{reward0}, Fig~\ref{reward1}, Fig~\ref{reward2}, Fig~\ref{reward3}, Fig~\ref{reward4}, Fig~\ref{reward5}, Fig~\ref{reward6} and Fig~\ref{reward7} are the reward for $UAV_{0}$, $UAV_{1}$, $UAV_{2}$, $UAV_{3}$, $UAV_{4}$, $UAV_{5}$, $UAV_{6}$, and $UAV_{7}$, respectively. When clustering unserved UEs, the threshold of distance equals the maximum coverage of UAV. And the initial position of UAV is random in their assigned area. Therefore some UAVs would happen to cover all UEs in their workplace while implementing BIRCH to assigning UAVs, such as $UAV_2$, $UAV_3$, $UAV_5$, and $UAV_7$. And other UAV, $UAV_0$, $UAV_1$, $UAV_4$,$UAV_6$, would cover all UEs easily, and the green curves would converge after steps of training because they are assigned to a suitable workplace where matches their maximum coverage.

\begin{figure}
    \centering
    \includegraphics[width=3.5in]{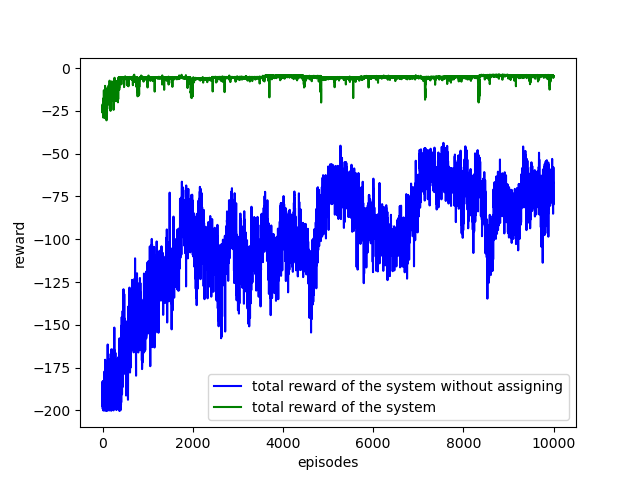}
    \caption{Total reward of the system}
    \label{fig:10}
\end{figure}

We mentioned the reward of the system \(r_{sys}\) in Section~\ref{pa}, and we plot it in Fig~\ref{fig:10}. It shows the system's reward per episode. As time goes by, both the green curve and blue curve rewards are increasing. But the green curve with assigning of BIRCH would obtain the higher reward stably, compared with the blue curve without assigning of BIRCH. After steps of learning, UAVs would complete the task to cover all unserved UEs with a reliable rate than the threshold we set before promptly.

\par
\begin{figure}
    \centering
    \includegraphics[width=3.5in]{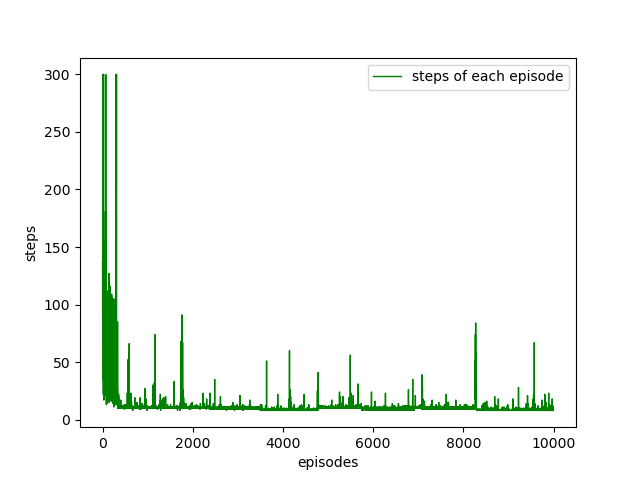}
    \caption{Steps per episode}
    \label{fig:9}
\end{figure}

\par
Besides, Figure~\ref{fig:9} shows the average number of steps that the system to cover all UEs per episode. Because the battery of UAVs we set is 300, steps below 300 mean that all UEs are covered in those episodes. After a period of training, the system could cover all users before the battery runs out. Besides, after the clustering with BIRCH and area dividing, we reduce the workload for each UAV. The steps for UAVs to cover all unconnected UEs decrease rapidly. UAVs would serve all UEs and complete the task after some steps of learning promptly.

\section{conclusion}
After the disaster, a stable wireless communication service is essential for rescuing; however, some TBSs are always crashed, and terrestrial emergency facilities cannot be deployed at once. With UAV development, we could deploy UAVs as aerial base stations because of UAV's agility. To cover maximum victims in disaster areas, we propose a Q-learning-based UAVs movement scheme cooperating with existing available TBSs to provide wireless communication for all victims. Before Q-learning, we use a clustering algorithm based on the coverage of aerial base stations to classify these unserved UEs and reduce the range of UAVs' workplaces, then deploy UAVs dynamically based on the result of clustering. Then UAVs could get the maximum rewards faster and easier in Q-learning. Through the simulation, UAVs could cover all users who are not served by TBSs post-disaster promptly.

\section*{Acknowledgment}
This work is partially supported by JSPS KAKENHI Grant Numbers JP19K20250, JP20F20080, and JP20H04174, Leading Initiative for Excellent Young Researchers (LEADER), MEXT, Japan, and JST, PRESTO Grant Number JPMJPR21P3, Japan. Mianxiong Dong is the corresponding author.

%

\begin{IEEEbiography}[{\includegraphics[width=1in,height=1.25in,clip,keepaspectratio]{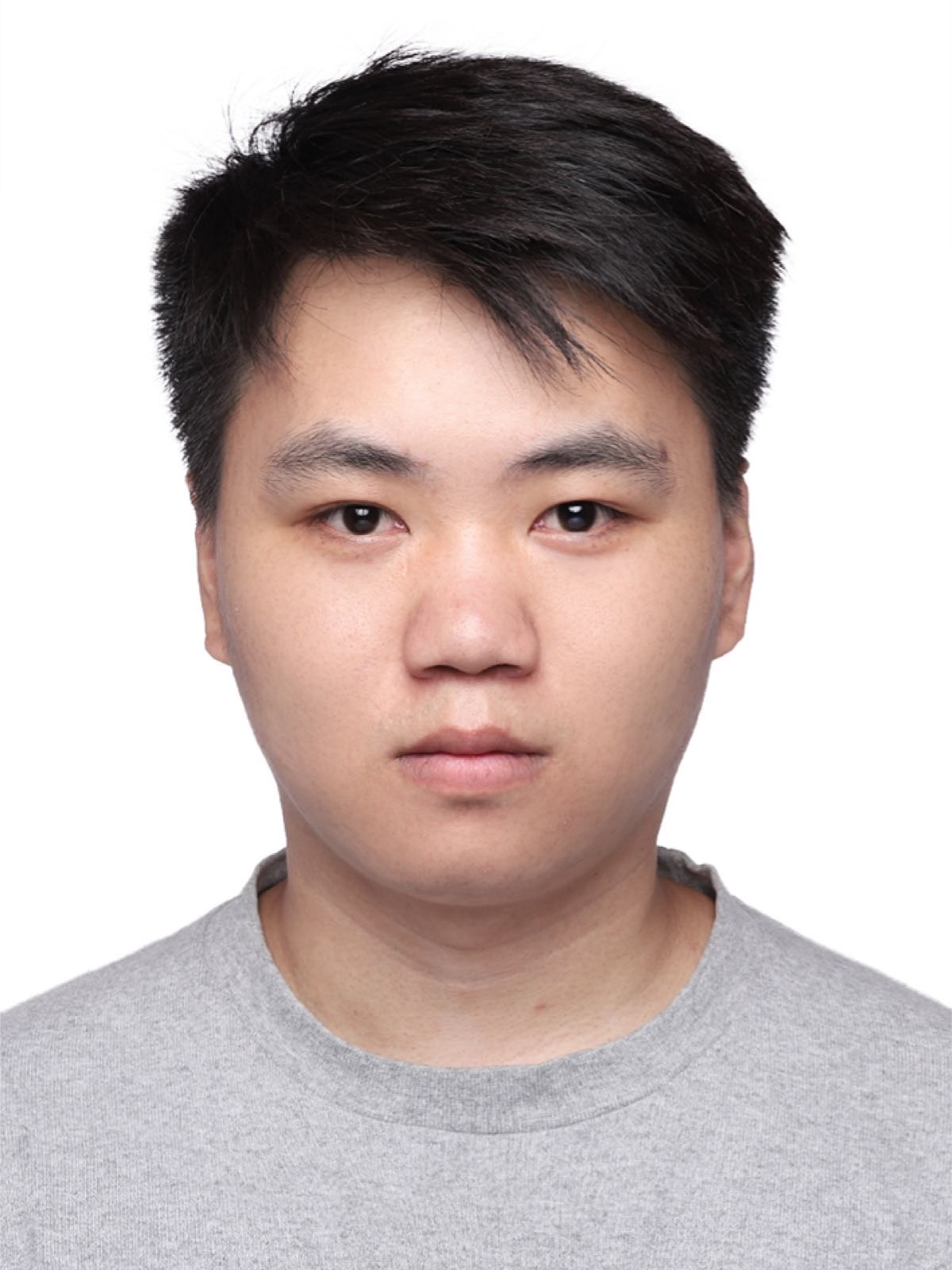}}]{Shiye Zhao}
    received the B.E. degree from Beijing University of Posts and Telecommunications, China, in 2019 and was a research student with ENeS lab at Muroran Institute of Technology, Japan, the same year. He is currently pursuing the M.S. degree in Muroran Institute of Technology, Japan.
\end{IEEEbiography}

\begin{IEEEbiography}[{\includegraphics[width=1in,height=1.25in,clip,keepaspectratio]{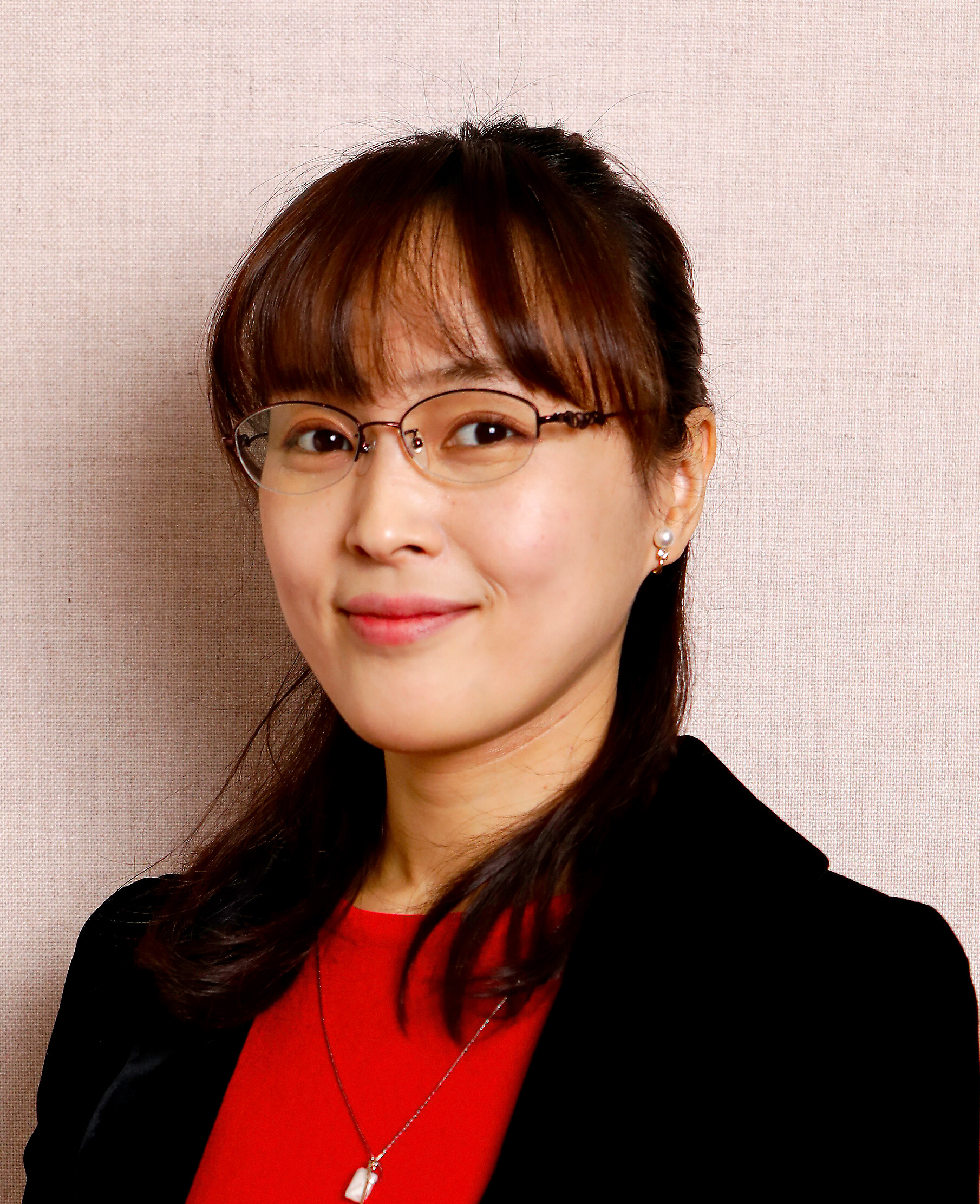}}]{Kaoru Ota}
	was born in Aizu-Wakamatsu, Japan. She received M.S. degree in Computer Science from Oklahoma State University, the USA in 2008, B.S. and Ph.D. degrees in Computer Science and Engineering from The University of Aizu, Japan in 2006, 2012, respectively. Kaoru is currently an Associate Professor and Ministry of Education, Culture, Sports, Science and Technology (MEXT) Excellent Young Researcher with the Department of Sciences and Informatics, Muroran Institute of Technology, Japan. From March 2010 to March 2011, she was a visiting scholar at the University of Waterloo, Canada. Also, she was a Japan Society of the Promotion of Science (JSPS) research fellow at Tohoku University, Japan from April 2012 to April 2013. Kaoru is the recipient of IEEE TCSC Early Career Award 2017, The 13th IEEE ComSoc Asia-Pacific Young Researcher Award 2018, 2020 N2Women: Rising Stars in Computer Networking and Communications, 2020 KDDI Foundation Encouragement Award, and 2021 IEEE Sapporo Young Professionals Best Researcher Award. She is Clarivate Analytics 2019, 2021 Highly Cited Researcher (Web of Science) and is selected as JST-PRESTO researcher in 2021, Fellow of EAJ in 2022.
\end{IEEEbiography}

\begin{IEEEbiography}[{\includegraphics[width=1in,height=1.25in,clip,keepaspectratio]{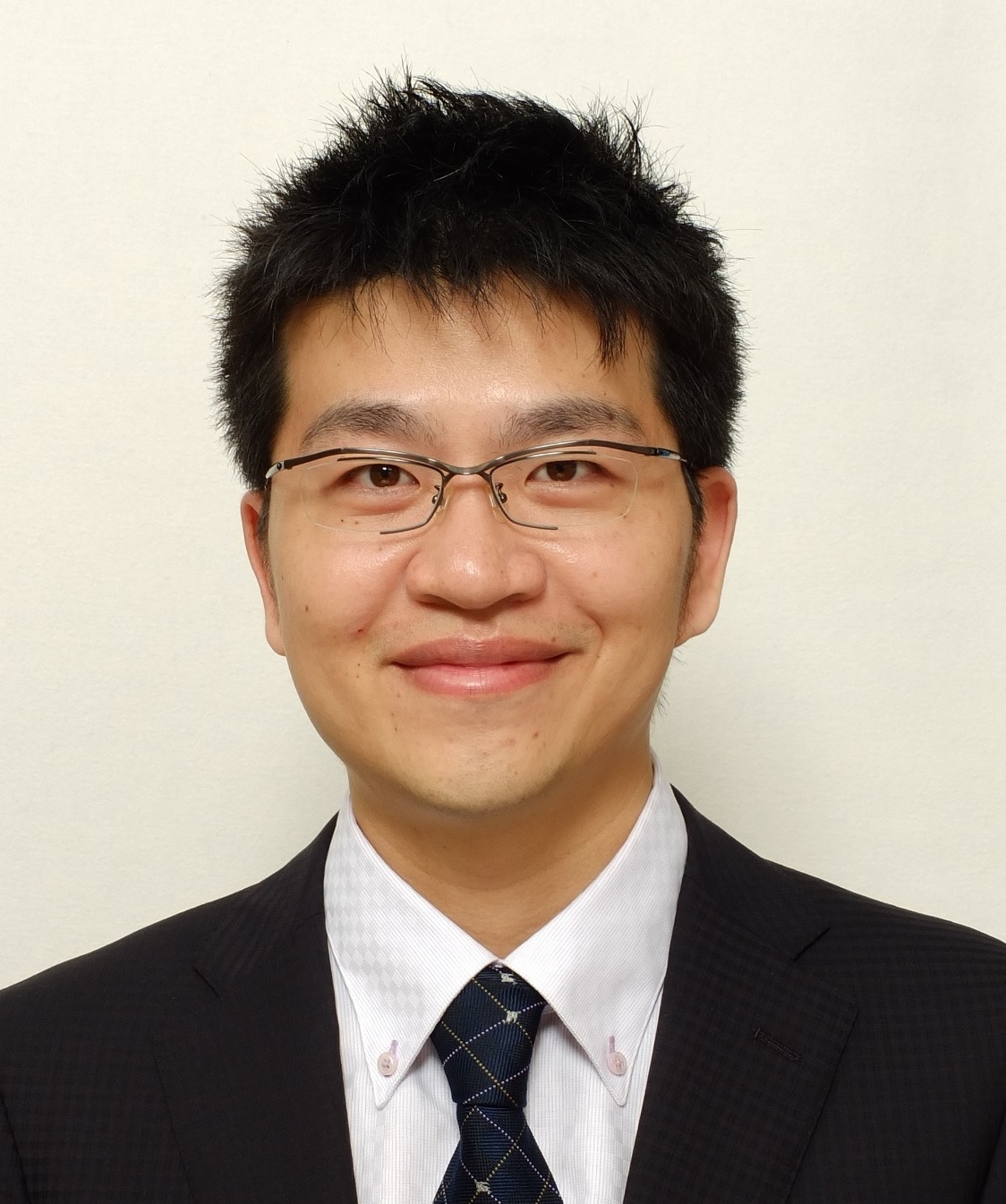}}]{Mianxiong Dong}
	received B.S., M.S. and Ph.D. in Computer Science and Engineering from The University of Aizu, Japan. He is the Vice President and Professor of Muroran Institute of Technology, Japan. He was a JSPS Research Fellow with School of Computer Science and Engineering, The University of Aizu, Japan and was a visiting scholar with BBCR group at the University of Waterloo, Canada supported by JSPS Excellent Young Researcher Overseas Visit Program from April 2010 to August 2011. Dr. Dong was selected as a Foreigner Research Fellow (a total of 3 recipients all over Japan) by NEC C\&C Foundation in 2011. He is the recipient of The 12th IEEE ComSoc Asia-Pacific Young Researcher Award 2017, Funai Research Award 2018, NISTEP Researcher 2018 (one of only 11 people in Japan) in recognition of significant contributions in science and technology, The Young Scientists’ Award from MEXT in 2021, SUEMATSU-Yasuharu Award from IEIEC in 2021, IEEE TCSC Middle Career Award in 2021. He is Clarivate Analytics 2019, 2021 Highly Cited Researcher (Web of Science) and Foreign Fellow of EAJ.
\end{IEEEbiography}




\end{document}